\pdfoutput=1

\documentclass[11pt]{article}

\usepackage[]{acl}

\usepackage{times}
\usepackage{latexsym}
\usepackage{multirow}
\usepackage{pgf}
\usepackage{tablefootnote}
\usepackage[T1]{fontenc}

\usepackage[utf8]{inputenc}

\usepackage{CJKutf8}

\usepackage{microtype}

\usepackage{tabularx}

\usepackage{graphicx}

\usepackage{hyperref}

\usepackage{multirow}
\usepackage[inline]{enumitem}
\usepackage{scrextend}
\usepackage{array}
\usepackage{arydshln}
\usepackage{graphicx}
\usepackage{hhline}

\usepackage{amsmath}

\usepackage{arydshln}
\title{SemEval-2022 Task 2: Multilingual Idiomaticity Detection and Sentence Embedding}

\author{
Harish Tayyar Madabushi\textsuperscript{1},
Edward Gow-Smith\textsuperscript{1}, \\
\textbf{Marcos Garcia}\textsuperscript{2},
\textbf{Carolina Scarton\textsuperscript{1}}, \\
\textbf{Marco Idiart\textsuperscript{3}} \and
\textbf{Aline Villavicencio\textsuperscript{1}} 
\\[0.3cm]
\textsuperscript{1} University of Sheffield, UK \\
\textsuperscript{2} Universidade de Santiago de Compostela, Spain \\
\textsuperscript{3} Federal University of Rio Grande do Sul, Brazil \\ 
\texttt{\small \{h.tayyarmadabushi, egow-smith1, c.scarton, a.villavicencio\} @sheffield.ac.uk} \\[-0.15cm]
\texttt{\small marcos.garcia.gonzalez@usc.gal, marco.idiart@gmail.com} 
\\
}

\begin{document}
\maketitle
\begin{abstract}
This paper presents the shared task on \textit{Multilingual Idiomaticity Detection and Sentence Embedding}, which consists of two Subtasks: (a) a binary classification task aimed at identifying whether a sentence contains an idiomatic expression, and (b) a task based on semantic text similarity which requires the model to adequately represent potentially idiomatic expressions in context. Each Subtask includes different settings regarding the amount of training data. Besides the task description, this paper introduces the datasets in English, Portuguese, and Galician and their annotation procedure, the evaluation metrics, and a summary of the participant systems and their results. The task had close to 100 registered participants organised into twenty five teams making over 650 and 150 submissions in the practice and evaluation phases respectively. 
\end{abstract}

\section{Introduction}
Multiword Expressions (MWEs) are a challenge for natural language processing (NLP), as their linguistic behaviour (e.g., syntactic, semantic) differs from that of generic word combinations \cite{baldwin-kim2010,ramisch-villavicencio2018}. Moreover, MWEs are pervasive in all domains \cite{biber99}, and it has been estimated that their size in a speaker’s lexicon of any language is of the same order of magnitude as the number of single words \cite{jackendoff97,erman-warren2000}, thus being of crucial interest for language modelling and for the computational representation of linguistic expressions in general.

One distinctive aspect of MWEs is that they fall on a \textit{continuum} of idiomaticity \cite{sag2002multiword,fazly-etal-2009-unsupervised,king-cook-2017-supervised}, as their meaning may or may not be inferred from one of their constituents (e.g., \textit{research project} being a type of `project', vs. \textit{brass ring} meaning a `prize'). In this regard, obtaining a semantic representation of a sentence which contains potentially idiomatic expressions involves both the correct identification of the MWE itself, and an adequate representation of the meaning of that expression in that particular context. As an example, it is expected that the representation of the expression \textit{big fish} will be similar to that of \textit{important person} in an idiomatic context, but closer to the representation of \textit{large fish} when conveying its literal meaning.

Classic approaches to representing MWEs obtain a compositional vector by combining the representations of their constituent words, but these operations tend to perform worse for the idiomatic cases. In fact, it has been shown that the degree of idiomaticity of a MWE can be estimated by measuring the distance between a compositional vector (obtained from the vectors of its components) and a single representation learnt from the distribution of the MWE in a large corpus \cite{cordeiro-etal-2019-unsupervised}.

Recent approaches to identify and classify MWEs take advantage of the contextualised representations provided by neural language models. On the one hand, some studies suggest that pre-training based on masked language modeling does not properly encode idiomaticity in word representations \cite{nandakumar-etal-2019-well,garcia-etal-2021-probing,garcia-etal-2021-assessing}. However, as these embeddings encode contextual information, supervised approaches using these representations tend to obtain better results in different tasks dealing with (non-)compositional semantics \cite{shwartz-dagan-2019-still,fakharian-cook-2021-contextualized,zeng-bhat-2021-idiomatic}.

As such, this shared task\footnote{Task website:~\href{https://sites.google.com/view/semeval2022task2-idiomaticity}{https://sites.google.com/view/semeval2022task2\-idiomaticity}}\textsuperscript{,}\footnote{GitHub:\href{https://github.com/H-TayyarMadabushi/SemEval_2022_Task2-idiomaticity}{https://github.com/H\-TayyarMadabushi/SemEval\_2022\_Task2-idiomaticity}} presents two Subtasks: i) Subtask A, to test a language model’s ability to detect idiom usage, and ii) Subtask B, to test the effectiveness of a model in generating representations of sentences containing idioms. Each of these Subtasks are further presented in two \emph{settings}: Subtask A in the Zero Shot and One Shot settings so as to evaluate models on their ability to detect previously unseen MWEs, and Subtask B in the Pre Train and the Fine Tune settings to evaluate models on their ability to capture idiomaticity both in the absence and presence of training data. Additionally, we provide strong baselines based on pre-trained transformer-based language models and release our codetr which participants can build upon.

\section{Related Tasks}
The computational treatment of MWEs has been of particular interest for the NLP community, and several shared tasks with different objectives and resources have been carried out.

The SIGLEX-MWE Section\footnote{\url{https://multiword.org/}} has organised various shared tasks, starting with the exploratory \textit{Ranking MWE Candidates} competition at the MWE 2008 Workshop, aimed at ranking MWE candidates in English, German and Czech.\footnote{\url{http://multiword.sourceforge.net/mwe2008}} More recently, together with the PARSEME community, they have conducted three editions of a shared task on the automatic identification of verbal MWEs \cite{savary-etal-2017-parseme,ramisch-etal-2018-edition,ramisch-etal-2020-edition}. In these cases, the objective is to identify both known and unseen verb-based MWEs in running text and to classify them under a set of predefined categories. Interestingly, these PARSEME shared tasks provide annotation guidelines and corpora for 14 languages, and include 6 categories (with additional subclasses) of verbal MWEs.

The \textit{Detecting Minimal Semantic Units and their Meanings (DiMSUM 2016)} shared task \cite{schneider-etal-2016-semeval} consisted of the identification of \textit{minimal semantic units} (including MWEs) in English, and labelling some of them according to a set of semantic classes (supersenses).

Focused on the interpretation of noun compounds, the \textit{Free Paraphrases of Noun Compounds} shared task of SemEval 2013 \cite{hendrickx-etal-2013-semeval} proposed to generate a set of free paraphrases of English compounds. The paraphrases should be ranked by the participants, and the evaluation is performed comparing these ranks against a list of paraphrases provided by human annotators.

Similarly, the objective of the SemEval 2010 shared task on \textit{The Interpretation of Noun Compounds Using Paraphrasing Verbs and Prepositions} \cite{butnariu-etal-2010-semeval} was to rank verbs and prepositions which may paraphrase a noun compound adequately in English (e.g., \textit{olive oil} as `oil \textit{extracted from} olive', or \textit{flu shot} as `shot \textit{to prevent} flu').

Apart from these competitions, various studies have addressed different tasks on MWEs and their compositionality, such as: classifying verb-particle constructions \cite{cook-stevenson-2006-classifying}, identifying light verb constructions and determining the literality of noun compounds \cite{shwartz-dagan-2019-still}, identifying and classifying idioms in running text \cite{zeng-bhat-2021-idiomatic}, as well as predicting the compositionality of several types of MWEs \cite{lin-1999-automatic,mccarthy-etal-2003-detecting,reddy-etal-2011-empirical,schulte-im-walde-etal-2013-exploring,salehi-etal-2015-word}.

\section{Dataset Creation}
The dataset used in this task extends that introduced by \citet{madabushi2021astitchinlanguagemodels}, also including Galician data along with Portuguese and English. Here we describe the four step process used in creating this dataset. 

The first step was to compile a list of 50 MWEs across the three languages. We sourced the MWEs in English and Portuguese from the Noun Compound Senses dataset (consisting of adjective-noun or noun-noun compounds) \cite{garcia-etal-2021-probing}, which extends the dataset by \citet{reddy-etal-2011-empirical} and provides human-judgements for compositionality on a Likert scale from 0 (non-literal/idiomatic) to 5 (literal/compositional). To ensure that the test set is representative of different levels compositionality, we pick approximately 10 idioms at each level of compositionality (0-1, 1-2, \dots). For Galician, we extracted noun-adjective compounds from the Wikipedia and the CC-100 corpora \cite{wenzek-etal-2020-ccnet} using the following procedure: First, we identified those candidates with at least 50 occurrences in the corpus. They were randomly sorted, and a native speaker and language expert of Galician selected 50 compounds from the list. The language expert was asked to take into account both the compositionality of the compounds (including idiomatic, partly idiomatic, and literal expressions), and their ambiguity (trying to select potentially idiomatic examples, i.e. compounds which can be literal or idiomatic depending on the context). 

In the second step of the dataset creation process, in English and Portuguese, annotators were instructed to obtain between 7 and 10 examples for each possible meaning of each MWE from news stories available on the web, thus giving between 20 and 30 total examples for each MWE. Each example consisted of three sentences: the target sentence containing the MWE and the two adjacent sentences. Annotators where explicitly instructed to select high quality examples, where neither of the two adjacent sentences were empty and, preferably, from the same paragraph. They were additionally required to flag examples containing novel meanings, so such new meanings of MWEs could be incorporated into the dataset. Sentences containing MWEs in Galician were directly obtained from the Wikipedia and the CC-100 corpora due to the sparsity of Galician data on the web. During this annotation step, we follow the method introduced by \citet{madabushi2021astitchinlanguagemodels}, and add two additional labels: `Proper Noun' and `Meta Usage'. `Meta Usage' represents cases wherein a MWE is used literally, but within a metaphor (e.g. \emph{life vest} in ``Let the Word of God be our life vest to keep us afloat, so as not to drown.'').

In the third phase, across all three languages, each possible meaning of each MWE was assigned a paraphrase by a language expert. For example, the compositional MWE \emph{mailing list} had the associated paraphrase `address list' added, whereas the idiomatic MWE \emph{elbow room} had the associated paraphrases `joint room', `freedom' and `space' added to correspond to each of its possible meanings. Language experts focused on ensuring that these paraphrases were as short as possible, so the resultant adversarial paraphrases could be used to evaluate the extent to which models capture nuanced differences in each of the meanings. 

The final phase of the process involved the annotation of each example with the correct paraphrase of the relevant MWE. This was carried out by two annotators, and any disagreements were discussed (in the case of Galician, in the presence of a language expert) and cases where annotators were not able to agree were discarded. 

\subsection{The Competition Dataset}

We use the training and development splits from \citet{madabushi2021astitchinlanguagemodels} with the addition of Galician data, and use the test split released by them as the evaluation split during the initial practice phase of the competition. We create an independent test set consisting of examples with new MWEs, and this set was used to determine the teams' final rankings. The labels for the evaluation and test sets are not released. We note that the competition is still active (in the `post-evaluation' phase), and open for submissions from anyone\footnote{\url{https://competitions.codalab.org/competitions/34710}}.

Since one of the goals of this task is to measure the ability of models to perform on previously unseen MWEs (Zero Shot) and on those for which they have very little training data (One Shot), we extract, where available, exactly one idiomatic and one compositional example associated with each MWE in the test data, which is released as associated One Shot training data. 

The final dataset consisted of 8,683 entries and the breakdown of the dataset is shown in \autoref{table:dataset-breakdown}. For further details on the training, development and practice evaluation splits, we direct readers to the work by \citet{madabushi2021astitchinlanguagemodels}. It should be noted that this original dataset does not contain data from Galician and so the only training data available in Galician was the One Shot training data. This was to evaluate the ability of models to transfer their learning across languages, especially to one that is low resourced. 

\begin{table}[ht]
\footnotesize
\def\arraystretch{1.1}
\centering
\begin{tabular}{|l|c|c|c|c|}
\hline
 & \multicolumn{3}{c|}{Language} & \\
\cline{2-4} 
Split & English & Portuguese & Galician & All\\
\hline
train& 3487 & 1290 & 63 & 4840 \\
dev & 466 & 273 & 0 & 739 \\
eval & 483 & 279 & 0 & 762 \\
test & 916 & 713 & 713 & 2342 \\
\hline
All & 5352 & 2555 & 776 & 8683 \\
\hline
\end{tabular}
\caption{\label{table:dataset-breakdown} Breakdown of the full dataset by language and data split.}
\end{table}

\section{Task Description and Evaluation Metrics}
SemEval-2022 Task 2 aims to stimulate research into a difficult area of NLP, that of handling non-compositional, or idiomatic, expressions. Since this is an area of difficulty for existing language models, we introduce two Subtasks; the first Subtask relates to idiomaticity detection, whilst the second relates to idiomaticity representation, success in which will require models to correctly encode idiomaticity. It is hoped that these tasks will motivate the development of language models better able to handle idiomaticity. Since we wish to promote multilingual models, we require all participants to submit results across all three languages. Both Subtasks are available in two settings, and participants are given the flexibility to choose which settings they wish to take part in. 

\begin{table*}[!htbp]
\footnotesize
\def\arraystretch{1.1}
\centering
\begin{tabular}{|l|l|p{10.5cm}|l|}
\hline
Language & MWE & Sentence & Label\\
\hline
English & old hat & Serve our favorite bourbon whiskeys in an \textbf{old hat} and we’d still probably take a sip or two. & 1 \\
English & old hat & But not all of the accouterments of power are \textbf{old hat} for the president. & 0 \\
Portuguese & força bruta & \textbf{Força Bruta} vai reunir alguns dos homens mais fortes do mundo. & 1 \\
Portuguese & força bruta & Gardner é conhecido por ser impulsivo e usar os poderes com grande impacto, de forma instintiva, com \textbf{força bruta}. & 0 \\
Galician & porta grande & Á esquerda da \textbf{porta grande}, en terra, observamos a tumba de “Don Manuel López Vizcaíno. & 1 \\
Galician & porta grande & Os dous dominadores da Copa Galicia 2017 regresaron pola \textbf{porta grande} ao certame autonómico na súa quinta xornada. & 0 \\

\hline
\end{tabular}
\caption{\label{table:subtask-a-examples} Examples for Subtask A. Note that the label 1 is assigned to non-idiomatic usage, which includes proper nouns, as in the Portuguese example.}
\end{table*}

\subsection{Subtask A: Idiomaticity Detection}
The first Subtask is a binary classification task, where sentences must be correctly classified into `idiomatic' (including `Meta Usage') or `non-idiomatic' / literal (including `Proper Noun'). Each example consists of the target sentence and two context sentences (sourced from either side of the target sentence) along with the relevant MWE. Some examples from this Subtask are shown in \autoref{table:subtask-a-examples}.

This Subtask is available in two settings: Zero Shot and One Shot. In the Zero Shot setting, the MWEs in the training set are disjoint from those in the development and test sets. Success in this setting will require models to generalise to unseen MWEs at inference time. In the One Shot setting, we include in the training set one idiomatic and one non-idiomatic example for each MWE in the development and test sets. This breakdown is shown in \autoref{table:train-dataset-breakdown}.

We use macro F1 score between the gold labels and predictions as the evaluation metric for this Subtask, due to the imbalanced datasets. 

\begin{table}[ht]
\footnotesize
\def\arraystretch{1.1}
\centering
\scalebox{0.85}{
\begin{tabular}{|l|c|c|c|c|c|}
\hline
 & & \multicolumn{3}{c|}{Language} & \\
\cline{3-5} 
Train Split & MWEs & English & Portuguese & Galician & All\\
\hline
Zero Shot & 236 & 3327 & 1164 & 0 & 4491\\
One Shot & 250 & 160 & 126 & 63 & 349 \\
\hline
Total & 486 & 3487 & 1290 & 63 & 4840\\
\hline
\end{tabular}
}
\caption{\label{table:train-dataset-breakdown} Breakdown of the training data into zero shot and one shot. Note that the MWEs in the zero shot and one shot data are disjoint.}
\end{table}

\subsection{Subtask B: Idiomaticity Representation}
The second Subtask is a novel idiomatic semantic textual similarity (STS) task, introduced by \citet{madabushi2021astitchinlanguagemodels}, where, given two input sentences, models must return an STS score between $0$ (least similar) and $1$ (most similar), indicating the similarity of the sentences. This requires models to correctly encode the meaning of non-compositional MWEs (idioms) such that the encoding of a sentence containing an idiomatic phrase (e.g. ``I initially feared that taking it would make me a \textbf{guinea pig}.'') and the same sentence with the idiomatic phrase replaced by a (literal) paraphrase (e.g. ``I initially feared that taking it would make me a \textbf{test subject}.'') are semantically similar to each other. Notice also that these two sentences, which mean the same thing, must necessarily be equally similar to any other third sentence. We choose this third sentence to be the sentence with the idiomatic phrase replaced by an \emph{incorrect} literal paraphrase (e.g. ``I initially feared that taking it would make me a \textbf{pig}.''). Such a sentence is the ideal adversarial example, and ensures that we test if models are making use of an incorrect meaning of the MWE in constructing a sentence representation.  

Data for this Subtask is generated in the following manner: MWEs in sentences are replaced by the literal paraphrase of one of its associated meanings. For example, the MWE `guinea pig' in the sentence ``I initially feared that taking it would make me a \textit{guinea pig}.'' is replaced by one of the literal paraphrases `test subject' or `pig' (see \autoref{table:subtask-b-examples}). Crucially, these replacements can either be with the correct paraphrase, or one that is incorrect. As such, there are two cases: \begin{itemize}
    \item The MWE has been replaced by its correct paraphrase. In this case, the similarity should be 1.\\
    $ sim(E, E_{\rightarrow \text{c}}) = 1 $
    \item The MWE has been replaced by its incorrect paraphrase. In this case, we require the model to give equivalent semantic similarities between this and the sentence where the MWE has been replaced by its correct paraphrase, and this and the original sentence.\\
    $ sim(E, E_{\rightarrow \text{i}}) = sim(E_{\rightarrow \text{c}}, E_{\rightarrow \text{i}}) $
\end{itemize}

Importantly, the task requires models to be \emph{consistent}. Concretely, the STS score for the similarity between a sentence containing an idiomatic MWE and that same sentence with the MWE replaced by the correct paraphrase must be equal to \emph{one} as this would imply that the model has correctly interpreted the meaning of the MWE. In the case where we consider the incorrect paraphrase, we check for consistency by requiring that the STS between the sentence containing the MWE and a sentence where the MWE is replaced by the incorrect paraphrase is equal to the STS between the sentence where the MWE is replaced by the correct paraphrase and one where it is replaced by the incorrect one. Notice, that all this does, is to require the model to, once again, interpret the meaning of the MWE to be the same (or very similar) to the correct literal paraphrase of that MWE. More formally, we require models to output STS scores for each example $E$ such that: 

\begin{equation}
\label{equation:task2}
\begin{split}
    \forall_{i \in I} \Big(  &sim(E, E_{\rightarrow \text{c}}) = 1; \\
    & sim(E, E_{\rightarrow \text{i}}) = sim(E_{\rightarrow \text{c}}, E_{\rightarrow \text{i}}) \Big)
\end{split}
\end{equation}

\noindent In Equation \ref{equation:task2} above, $E_{\rightarrow \text{c}}$ represents an example containing the MWE $E$, wherein that MWE is replaced by its \emph{correct} contextual paraphrase. $E_{\rightarrow \text{i}}$ on the other hand, represents the example wherein the MWE $E$ is replaced by one of its \emph{incorrect} contextual paraphrases. Examples for this Subtask are shown in \autoref{table:subtask-b-examples}.

\begin{table*}[!htbp]
\footnotesize
\def\arraystretch{1.1}
\centering
\begin{tabular}{|p{3.8cm}|p{3.8cm}|p{3.8cm}|p{2.9cm}|}
\hline
Sentence (E) & Correct Replacement (\scalebox{0.8}{$E_{\text{MWE} \rightarrow \text{c}}$}) & Wrong Replacement (\scalebox{0.8}{$ E_{ \text{MWE}\rightarrow \text{i}}$}) & Expected \\
\hline
And finally, the snow falls again, this time in a thick, \textbf{wet blanket} that encapsulates everything. & And finally, the snow falls again, this time in a thick, \textbf{damp blanket} that encapsulates everything. & And finally, the snow falls again, this time in a thick, \textbf{killjoy} that encapsulates everything. & \scalebox{0.7}{ 
\begin{tabular}{@{}l@{}}
  $sim(E, E_{\rightarrow \text{c}}) = 1$ \\ 
  $sim(E, E_{\rightarrow \text{i}}) = sim(E_{\rightarrow \text{c}}, E_{\rightarrow \text{i}})$
 \end{tabular}}\\
\hdashline
I initially feared that taking it would make me a \textbf{guinea pig}. & I initially feared that taking it would make me a \textbf{test subject}. &
I initially feared that taking it would make me a \textbf{pig}. &  \scalebox{0.7}{ 
\begin{tabular}{@{}l@{}}
  $sim(E, E_{\rightarrow \text{c}}) = 1$ \\ 
  $sim(E, E_{\rightarrow \text{i}}) = sim(E_{\rightarrow \text{c}}, E_{\rightarrow \text{i}})$
 \end{tabular}}\\
\hline
\end{tabular}
\caption{\label{table:subtask-b-examples} Examples for Subtask B. For brevity we only include examples in English.}
\end{table*}

Since this task relies on models' ability to correctly assign STS scores for sentences with do not contain idiomatic MWEs, we additionally include standard STS data in our test data. This has the added benefit of preventing models from overfitting on the MWE dataset. We include this STS evaluation data from the STS Benchmark dataset \cite{cer2017semeval} in English and the ASSIN2 STS dataset \cite{real2020assin} in Portuguese. There is no available STS data for Galician, so none is included. We use the Spearman's rank correlation coefficient between the two sets of STS scores generated by models as the evaluation metric in this Subtask. We do not use Pearson correlation as it has been shown to be a poor indicator of performance on STS tasks \cite{reimers2016task}. 

This Subtask is also available in two settings: the Pre Train setting and the Fine Tune setting. In the Pre Train setting, we require that models are \emph{not} trained on idiomatic STS data. However, models can be trained (including ``fine-tuned'') on any other training objective (such as during the pre-training of language models). The Fine Tune setting, on the other hand, allows all training regimes, including the fine-tuning on any idiomatic STS dataset. 

\subsection{Baselines}
In order to generate baseline results, we used pre-trained transformer-based \cite{vaswani2017attention} language models. We use multilingual BERT \cite{devlin2018bert} to benefit from cross-lingual transfer. For both settings in Subtask A, we simply Fine Tune the pre-trained model on the training data provided. For the Zero Shot setting, we include the context sentences, whereas in the One Shot setting, we exclude the context sentences but add the MWE as a second sentence. This is based on the best-performing approaches found by \citet{madabushi2021astitchinlanguagemodels}. 

For Subtask B Pre Train, we introduce single tokens for each MWE in the data. This is motivated by the `idiom principle' \cite{sinclair1991corpus}, which hypothesises that humans process idioms by treating them as a single unit. Since BERT embeddings cannot be directly used for STS, we create a sentence transformer model \cite{reimers-2019-sentence-bert} using multilingual BERT with these added tokens, and train it on the English and Portuguese STS data. Importantly, the new tokens introduced for MWEs are randomly initialised and no continued pre-training is performed. As such, they serve to `break compositionality' rather than to create more effective representations of MWEs. This breaking of compositionality has been shown to be effective by \citet{madabushi2021astitchinlanguagemodels}.

For the Fine Tune setting, the same approach is taken, although no training is done on the STS data, and instead we Fine Tune on the training data provided. This lack of training on the STS data is intentional as we intend to establish the effectiveness of the MWE based training data, and are reflected by the comparatively lower scores on the STS subsection of the test data (\autoref{table:subtask-b-fine-tune-results}). 

It should be noted that these baseline methods that make use of multilingual BERT are particularly strong when compared to typical `baselines'. This is intentional as we aim to promote the development of models that are comparable to the current state-of-the-art.

\section{Participating Systems and Results}
Twenty five teams in total participated, with the most participants to Subtask A Zero Shot (20). The results for the individual Subtasks are given in \autoref{table:subtask-a-zero-shot-results}, \autoref{table:subtask-a-one-shot-results}, \autoref{table:subtask-b-pre-train-results} and \autoref{table:subtask-b-fine-tune-results}. Here we discuss the methods used by the best-performing teams as well as some interesting approaches. Full details of methods used by participants is given in \autoref{app:method-breakdown}.

\begin{CJK*}{UTF8}{gbsn}
\begin{table*}[ht!]
\footnotesize
\def\arraystretch{1.1}
\centering
\begin{tabular}{|c|l|c|c|c|c|}
\hline
& &\multicolumn{3}{c|}{Language} &   \\
\cline{3-5}
Ranking & Team & English & Portuguese & Galician & All\\
\hline
1&clay&  0.9016& 0.8277& 0.9278& 0.8895 \\
2&yxb&  0.8948& 0.8395& 0.7524& 0.8498\\
3&NER4ID \cite{ner4id-2022-semeval}&  0.8680& 0.7039& 0.6550& 0.7740\\
4&HIT \cite{hit-2022-semeval}& 0.8242& 0.7591& 0.6866& 0.7715\\
5&Hitachi \cite{hitachi-2022-semeval}& 0.7827& 0.7607& 0.6631& 0.7466\\
6&OCHADAI \cite{ochadai-2022-semeval}& 0.7865& 0.7700& 0.6518& 0.7457\\
7&yjs& 0.8253& 0.7424& 0.6020& 0.7409\\
8&CardiffNLP-metaphors \cite{cardiffnlp-2022-semeval}& 0.7637& 0.7619& 0.6591& 0.7378\\
9&Mirs& 0.7663& 0.7617& 0.6429& 0.7338\\
10&Amobee& 0.7597& 0.7147& 0.6768& 0.7250\\
11&HYU \cite{hyu-2022-semeval}& 0.7642& 0.7282& 0.6293& 0.7227\\
12&Zhichun Road \cite{zhichunroad-2022-semeval}& 0.7489& 0.6901& 0.5104& 0.6831\\
13&海鲛NLP& 0.7564& 0.6933& 0.5108& 0.6776\\
14&UAlberta \cite{ualberta-2022-semeval}& 0.7099& 0.6558& 0.5646& 0.6647\\
15&Helsinki-NLP \cite{helsinki-2022-semeval}& 0.7523& 0.6939& 0.4987& 0.6625\\
16&daminglu123 \cite{daminglu-2022-semeval}& 0.7070& 0.6803& 0.5065& 0.6540\\
\hdashline
 &baseline \cite{madabushi2021astitchinlanguagemodels}& 0.7070 & 0.6803 & 0.5065 & 0.6540\\
\hdashline
17&kpfriends \cite{kpfriends-2022-semeval}& 0.7256& 0.6739& 0.4918& 0.6488\\
18&Unimelb\textunderscore AIP& 0.7614& 0.6251& 0.5020& 0.6436\\
19&YNU-HPCC \cite{ynu-hpcc-2022-semeval}& 0.7063& 0.6509& 0.4805& 0.6369\\
20&Ryan Wang& 0.5972& 0.4943& 0.4608& 0.5331\\
\hline
N/A&JARVix \cite{jarvix-2022-semeval}\footnotemark & 0.7869 & 0.7201 & 0.5588 & 0.7235 \\
\hline
\end{tabular}
\caption{\label{table:subtask-a-zero-shot-results} Results for Subtask A Zero Shot. The evaluation metric is macro F1 score, and the ranking is based on the `All' column. }
\end{table*}
\end{CJK*}

\subsection{Subtask A Zero-Shot}
Of the twenty teams that submitted to this setting, 12 reported using transformer-based approaches. The best-performing team (clay) used different masking strategies during pretraining, and performed finetuning with data augmentation (including back-translation, \citealp{edunov2018understanding}) as well as using soft-label finetuning (a knowledge distillation approach). The team in second (yxb) used a multilingual T5 model \cite{xue2020mt5} with various data augmentation techniques including: back-translation; synonym replacement; random insertion, swap, and deletion. They also used an alternative loss function for unbalanced data, called focal loss \cite{lin2017focal}.
The third team (NER4ID; \citealp{ner4id-2022-semeval}) used a dual-encoder architecture to encode the MWE and its context, then predicted idiomaticity by looking at the similarity score. This approach has a precedent in previous work that hypothesises the semantic similarity between a MWE and its context to be a good indicator of idiomaticity \cite{liu2018heuristically}. They also implemented named entity recognition as an intermediate step which they found provided great improvements. Interestingly, two teams (UAlberta; \citealp{ualberta-2022-semeval}, and Unimelb\textunderscore AIP) used unsupervised approaches, i.e. not using any of the provided training data. UAlberta were able to beat the baseline using translation information from resources such as Open Multilingual Wordnet \cite{bond2013linking} and BabelNet \cite{navigli2010babelnet}. They hypothesised that for idiomatic MWEs, the individual words are less likely to share mult-synsets with their translations. They also used a POS tagger for identifying proper nouns. 

\subsection{Subtask A One Shot}
The best-performing team (HIT; \citealp{hit-2022-semeval}) used XLM-R \cite{conneau2019unsupervised}, and added `[SEP]' tokens around the relevant MWE in the target sentence, unless it was capitalised, in which case they excluded these tokens. This is an alternative approach to that of \citet{madabushi2021astitchinlanguagemodels}, where the MWEs were added as a second sentence. They also used R-Drop \cite{wu2021r} as a regularisation method. The second best team (kpfriends; \citealp{kpfriends-2022-semeval}) used an ensemble of checkpoints with soft-voting. They also started with XLM-RoBERTa (large) trained on CoNLL. Interestingly, this team had the largest difference in performance across the two settings of Subtask A (coming in 16th in the Zero Shot setting).
The third best team (UAlberta; \citealp{ualberta-2022-semeval}) used a transformer-based classifier with additional features of glosses for the individual words of the relevant MWE. They hypothesised that this would help for determining compositionality, since the meaning of compositional MWEs could be deduced from the glosses of the individual words. 
An interesting approach was taken by MaChAmp \cite{machamp-2022-semeval}, who used multi-task learning across multiple SemEval tasks (2, 3, 4, 6, 10, 11, 12), pretraining a Rebalanced mBERT (RemBERT) \cite{chung2020rethinking} model across all of the tasks, then retraining a model for each specific task. Since for this task we do not allow the use of additional data, we do not include this team in the ranking, but their score is reported for reference.

\footnotetext[6]{Not ranked due to only submitting to the `post-evaluation' phase.}

\begin{CJK*}{UTF8}{gbsn}

\begin{table*}[ht]
\footnotesize
\def\arraystretch{1.1}
\centering
\begin{tabular}{|c|l|c|c|c|c|}
\hline
& &\multicolumn{3}{c|}{Language} &  \\
\cline{3-5} 
Ranking & Team & English & Portuguese & Galician &  All \\
\hline
1&HIT \cite{hit-2022-semeval}&0.9639& 0.8944& 0.9369& 0.9385 \\
2&kpfriends \cite{kpfriends-2022-semeval}& 0.9606& 0.8993& 0.9215& 0.9346 \\ 
3&UAlberta \cite{ualberta-2022-semeval}& 0.9453& 0.8918& 0.9120& 0.9243 \\
4&Zhichun Road \cite{zhichunroad-2022-semeval}& 0.9344& 0.8559& 0.8927& 0.9033 \\
5&clay& 0.9181& 0.8423& 0.9313& 0.9022 \\
6&YNU-HPCC \cite{ynu-hpcc-2022-semeval}& 0.9179& 0.8633& 0.8781& 0.8948 \\
7&CardiffNLP-metaphors \cite{cardiffnlp-2022-semeval}& 0.9464& 0.8385& 0.8545& 0.8934 \\
8&yxb& 0.8995& 0.8266& 0.8781& 0.8779 \\
9&NER4ID \cite{ner4id-2022-semeval}& 0.9079& 0.8179& 0.8695& 0.8771 \\
10&HYU \cite{hyu-2022-semeval}& 0.9159& 0.8457& 0.8287& 0.8750 \\ 
11&yjs& 0.9199& 0.8365& 0.8294& 0.8747 \\ 
\hdashline
&baseline \cite{madabushi2021astitchinlanguagemodels}& 0.8862 & 0.8637 & 0.8162 & 0.8646 \\
\hdashline
12&Mirs& 0.7570& 0.7549& 0.6712& 0.7367 \\ 
13&daminglu123 \cite{daminglu-2022-semeval}& 0.7486& 0.7085& 0.6004& 0.7040 \\
14&海鲛NLP& 0.7649& 0.7156& 0.5134& 0.6851 \\ 
15&OCHADAI \cite{ochadai-2022-semeval}& 0.7069& 0.6445& 0.5235& 0.6573 \\ 
16&Ryan Wang& 0.3314& 0.4058& 0.3779& 0.4044 \\ 
\hline
N/A&MaChAmp \cite{machamp-2022-semeval}\footnotemark& 0.7204& 0.6247& 0.5532& 0.6607 \\ 
N/A&JARVix \cite{jarvix-2022-semeval}\footnotemark& 0.8410 & 0.8162 & 0.7918 & 0.8243 \\
\hline
\end{tabular}
\caption{\label{table:subtask-a-one-shot-results} Results for Subtask A One Shot. The evaluation metric is macro F1 score, and the ranking is based on the `All' column.}
\end{table*}
\end{CJK*}

\subsection{Subtask B Pre Train}
No teams reported using non-transformer-based approaches for this setting.
The best-performing team (drsphelps; \citealp{drsphelps-2022-semeval}) used a modification of the baseline with BERT for Attentive Mimicking (BERTRAM) \cite{schick2019bertram} to generate embeddings as replacements for the randomly-initialised one token embeddings used by the baseline. This method takes both form and context into account, thus not assuming total non-compositionality as the one-token method does. 
It should be noted that every team in this setting improved upon the baseline result.

\begin{table*}[ht]
\footnotesize
\def\arraystretch{1.1}
\centering
\begin{tabular}{|c|l|c|c|c|}
\hline
& & \multicolumn{2}{c|}{Subset} &  \\
 \cline{3-4} 
Ranking & Team & Idiom Only & STS Only & All \\
\hline
1 & drsphelps \cite{drsphelps-2022-semeval}& 0.4030 & 0.8641 & 0.6402 \\
2 & colorful & 0.4290 & 0.8880 & 0.6262 \\
3 & Mirs & 0.3750 & 0.8623 & 0.6038 \\
4 & Zhichun Road \cite{zhichunroad-2022-semeval}& 0.2826 & 0.8359 & 0.5632 \\
5 & YNU-HPCC \cite{ynu-hpcc-2022-semeval}& 0.2872 & 0.7125 & 0.5577 \\
6 & ALTA & 0.2154 & 0.8608 & 0.5379 \\
\hdashline
 & baseline \cite{madabushi2021astitchinlanguagemodels}& 0.2263 & 0.8311 & 0.4810 \\
\hline
\end{tabular}
\caption{\label{table:subtask-b-pre-train-results} Results for Subtask B Pre Train. The evaluation metric is Spearman correlation, and the ranking is based on the `All' column.}
\end{table*}

\subsection{Subtask B Fine Tune}
No teams reported using non-transformer-based approaches for this setting.
The best-performing team (YNU-HPCC; \citealp{ynu-hpcc-2022-semeval}) used a pretrained Sentence-BERT \cite{reimers-2019-sentence-bert} model, then finetuned using multiple negatives ranking loss \cite{henderson2017efficient} and triplet loss.
The second best team (drsphelps; \citealp{drsphelps-2022-semeval}) used an identical approach to that in Subtask B Pre Train, using BERTRAM \cite{schick2019bertram}, with additional finetuning on the training data provided. 
The third best team (Eat Fish) used a multilingual model pretrained with knowledge distillation, as well as data augmentation.

\begin{table*}[ht]
\footnotesize
\def\arraystretch{1.1}
\centering
\begin{tabular}{|c|l|c|c|c|}
\hline
& & \multicolumn{2}{c|}{Subset} &   \\
 \cline{3-4} 
Ranking & Team & Idiom Only & STS Only & All \\
\hline
1 & YNU-HPCC \cite{ynu-hpcc-2022-semeval}& 0.4277 & 0.6637 & 0.6648 \\
2 & drsphelps \cite{drsphelps-2022-semeval}& 0.4124 & 0.8188 & 0.6504 \\
3 & Eat Fish & 0.3688 & 0.8660 & 0.6475 \\
4 & Zhichun Road \cite{zhichunroad-2022-semeval}& 0.3956 & 0.5615 & 0.6401 \\
\hdashline
 & baseline \cite{madabushi2021astitchinlanguagemodels}&  0.3990 & 0.5961 & 0.5951 \\
\hdashline
5 & ALTA & 0.2566 & 0.6156 & 0.5755 \\
\hline
\end{tabular}
\caption{\label{table:subtask-b-fine-tune-results} Results for Subtask B Fine Tune. The evaluation metric is Spearman correlation, and the ranking is based on the `All' column.}
\end{table*}

\subsection{Overview of Submissions}

\begin{figure}[ht!]
\includegraphics[width=\columnwidth]{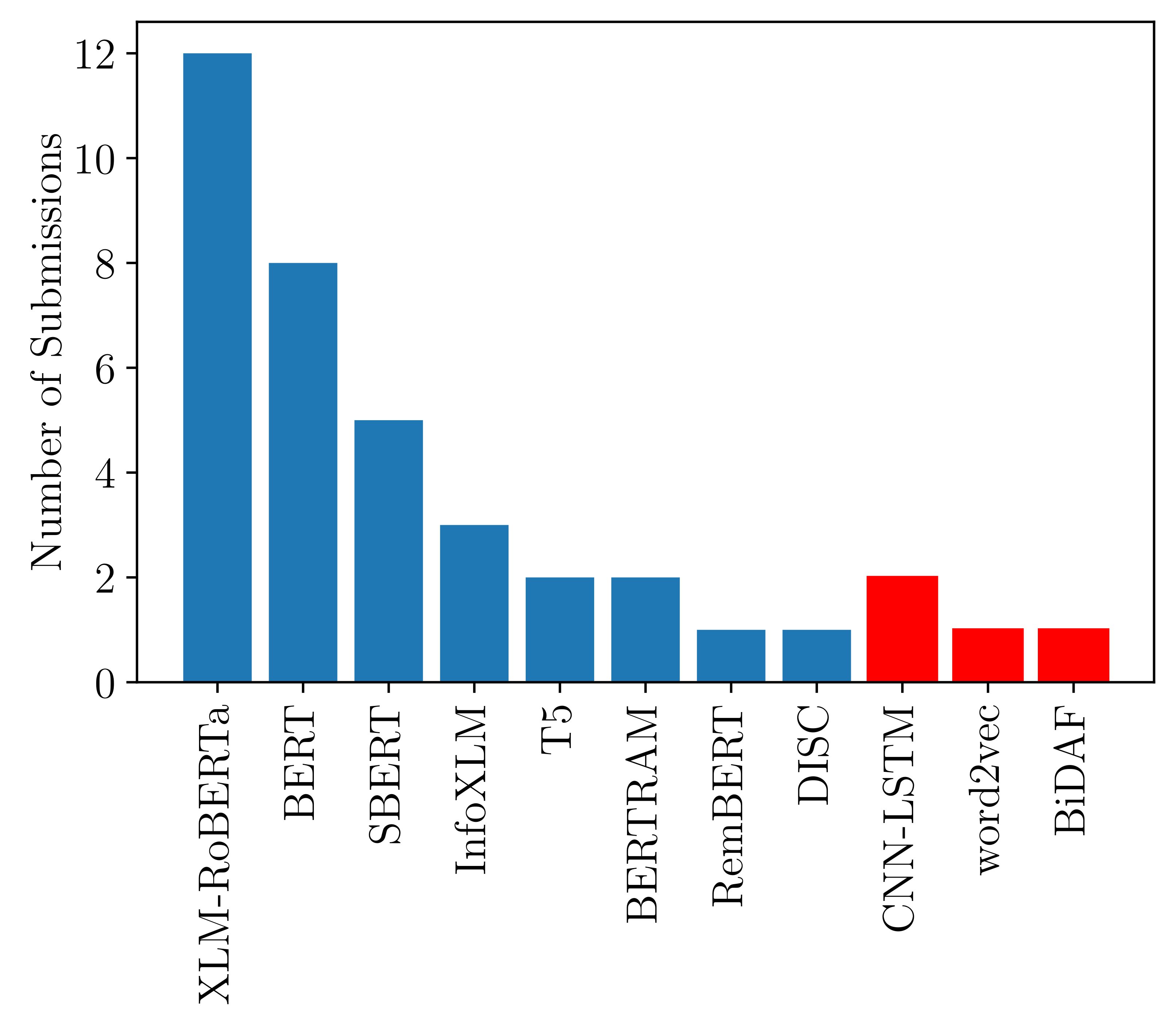}
\caption{\label{fig:num_sub_models} Models mentioned in the submissions. In blue are models that use transformers either wholly or partially, whilst in red are alternative models.}
\end{figure}

In \autoref{fig:num_sub_models} we show the models that were mentioned in the submissions.

The majority of participants used transformer-based approaches, although in both settings for Subtask A there were three teams using other approaches.
In Subtask B, as mentioned previously, no non-transformer approaches were mentioned, which is expected since this task was designed for the pretrain-finetune paradigm.

\begin{figure}[ht!]
\includegraphics[width=\columnwidth]{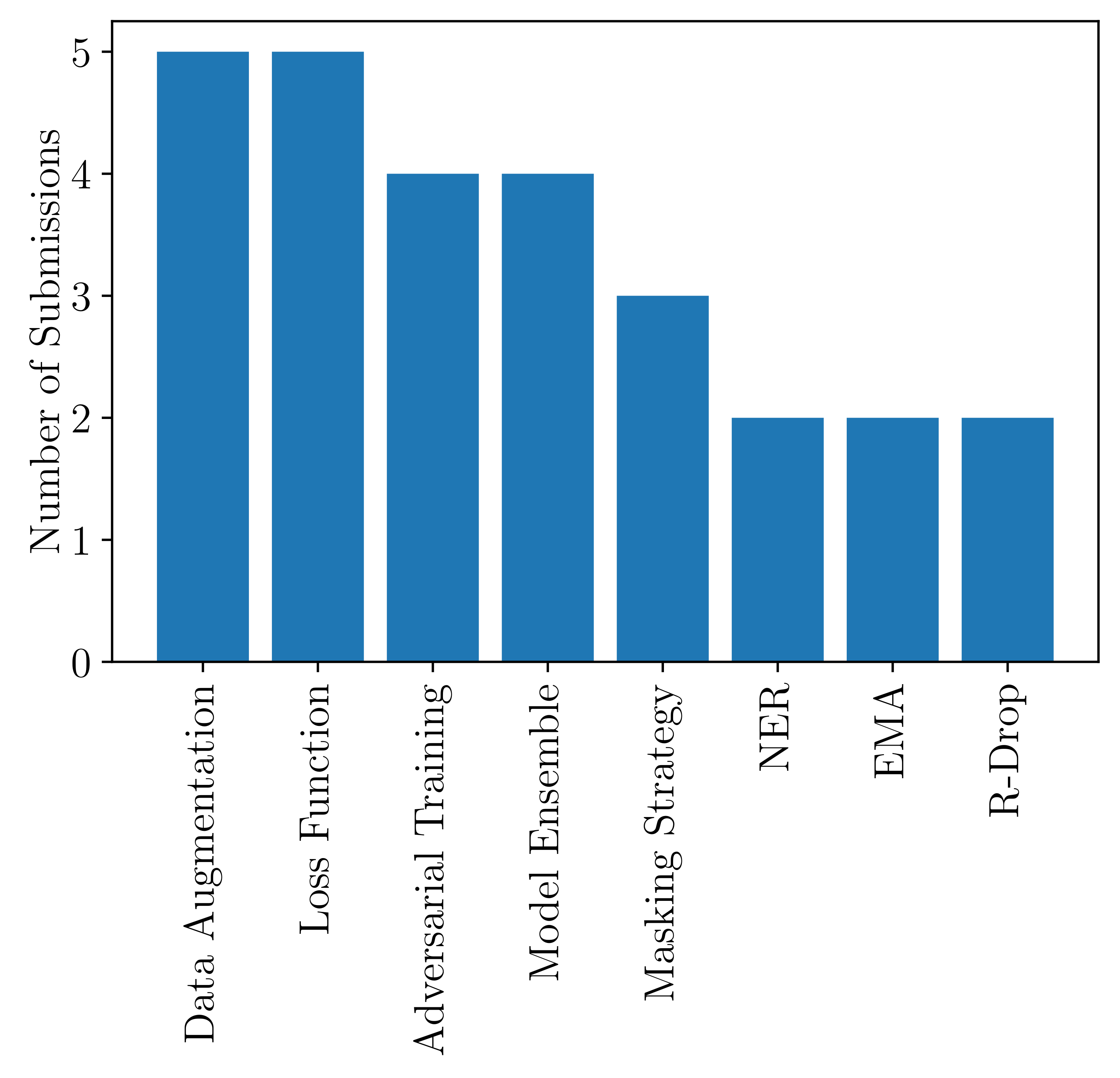}
\caption{\label{fig:num_sub_methods} Methods mentioned in more than one submission.}
\end{figure}

In \autoref{fig:num_sub_methods} we show the methods mentioned in more than one submission. Data augmentation approaches were popular, the most frequently-mentioned being back-translation \cite{edunov2018understanding}. Equally as popular were approaches using alternative loss functions.

\footnotetext[7]{Not ranked due to using a multi-task learning approach.}
\footnotetext[8]{Not ranked due to only submitting to the `post-evaluation' phase.}

\section{Methods}
The primary goal of this shared task was to provide a platform for the evaluation of a variety of methods for the identification and represention of MWEs. This section gives an overview of the methods that have been successful in each of the Subtasks. In particular, we attempt to identify the combination of methods across submissions that have significant potential for future development.

\subsection{Subtask A}
Subtask A, the identification of MWEs, comprised two settings: Zero Shot and One Shot. Crucially, the results from the task show that \emph{methods that are successful in the Zero Shot setting, fail to be successful in the One Shot setting and vice versa}. The two problems seem to require capabilities that are quite distinct. This seems intuitive when translated into the kind of thinking that one might use in identifying idioms: When one hears an idiom for the first time, we are likely to recognise that it sounds `idiom-like' based on our prior understanding of idioms, whereas when we come across an idiom that we are familiar with, we link our existing knowledge of that idiom with the current instance of it. 

This seems to play out in the successful models in this Subtask, as the general trend amongst the methods that were successful in the Zero Shot setting, with one exception, is the generalisation of models using regularisation, data augmentation or dropout. While regularisation did feature in the top performing model in the One Shot setting, it seems to have been less important to generalise models when they had access to as little as one training example associated with each model. The best performing linguistically motivated method -- which compares the semantic similarity between the MWE span and that of the surrounding context -- ranked third in the Zero Shot setting, although it performed 11 points below the best performing method. This is of particular interest as this method has previously been shown to be extremely powerful in detecting idiomaticity in non-contextual models.

Models successful in the One Shot setting, again with one exception, seem to be those which are more powerful at extracting cues from the minimal training examples and tended to be larger, ensembled or trained to a larger extent using adversarial training. The best performing method which incorporated elements based on linguistic theory also ranked third in this setting and incorporated the gloss of each individual word in the target MWE to aid in models' ability to detect compositionality. 

Interestingly, the use of the idiom principle in creating single token representations for MWEs is absent amongst the methods used for this Subtask. While such a comparison would have been interesting, it is hardly surprising that this method is not amongst those used, given that the cost of pre-training with new MWE tokens is rather high. 

\subsection{Subtask B}
Subtask B, the novel task of creating contextual representations of MWEs which are consistent with the paraphrased version of that MWE as measured by Spearman's rank correlation, coefficient also had two settings: the first without associated training examples (Pre Train) and the second with (Fine Tune). Since the sentence embeddings generated by pre-trained language models cannot be directly compared for similarity, such models must be altered so as to be used for this Subtask. Additionally, as pointed out by \citet{madabushi2021astitchinlanguagemodels}, the MWEs contained within sentences can be represented using single tokens even without pre-training, as the `breaking' of compositionality itself produces more accurate representations of sentences containing MWEs. 

As such, models that perform the best on the Pre Train setting focus on the creation of more accurate single token representations of MWEs, while the top performing models on the Fine Tune setting, in general, focus on optimising sentence similarity. This seems to be consistent with the observation by \citet{madabushi2021astitchinlanguagemodels} that fine-tuning is indeed a reasonable way of learning the representation of MWEs. It should be noted that these trends are less certain since there are fewer participants on this Subtask, some of whom do not share their methods, and the one team that we know used a method of learning new representations of MWEs is ranked first in the Pre Train setting but ranked second in the Fine Tune setting. 

\section{Conclusions and Future work}

We present, in this paper, `SemEval 2022 Task2: Multilingual Idiomaticity Detection and Sentence Embedding', consisting of two Subtasks: i) Subtask A, to test a language model’s ability to detect idiom usage, and ii) Subtask B, to test a model's ability to generate representations of sentences containing idioms. This task, aimed at boosting research into the detection and representation of idiomatic expressions, had submissions from 25 teams consisting of close to 100 participants. 

We additionally provide an overview and analysis of the methods used by participants, which we believe will help future research in this field. In particular, we highlight the need for distinct methods when detecting MWEs that have been previously seen and when detecting ones that have not. In representing idiomatic expressions, we show, through the novel idiomatic STS task presented here, that models are rather effective when they have training data available, but, as demonstrated in the Pre Train setting, more methods of encoding MWEs are required when training data is not available. 

While the top performing methods across this task have been driven by deep neural models independent of linguistic features, we highlight that this does not imply that the addition of linguistically motivated features does not lead to improvements on the task. Instead, it points to the possibility of integrating these methods into the more powerful neural models in future work where an ablation study might shed more light on the impact of each feature. 

\subsection*{Acknowledgements}

This work was partially supported by the UK EPSRC grant EP/T02450X/1, by the CDT in Speech and Language Technologies and their Applications (UKRI grant number EP/S023062/1), by a \textit{Ramón y Cajal} grant (RYC2019-028473-I), and by the grant ED431F 2021/01 (Galician Government).

\bibliography{anthology,custom}
\bibliographystyle{acl_natbib}

\clearpage
\appendix

\section{Full Breakdown of Methods}
\label{app:method-breakdown}

All participants were invited to submit a short description of their methods, as well as to submit a paper. In \autoref{table:subtask-a-zero-shot-methods}, \autoref{table:subtask-a-one-shot-methods}, \autoref{table:subtask-b-pre-train-methods}, and \autoref{table:subtask-b-fine-tune-methods} we give all the method descriptions that were submitted. 

\begin{CJK*}{UTF8}{gbsn}
\begin{table*}[ht]
\tiny
\def\arraystretch{1.1}
\centering
\begin{tabularx}{\linewidth}{|c|c|p{12.75cm}|}
\hline
Ranking & Team & Method  \\
\hline
1&clay& "domain pretraining with different masking strategies
finetuning with data augmentation such as back-translation
finetuning with soft label from former checkpoint" \\
2&yxb& "use mT5-Base
use Easy Data Augmentation techniques include back-translation, synonym replacement, random insertion, random swap, random deletion
include label unbalanced loss function：focal loss
use model ensemble" \\
3&NER4ID& "Dual-encoder (Transformer-based) architecture that encodes both the potentially idiomatic expression and its context, and predicts idiomaticity by looking at their similarity score: high similarity -> compositional, low similarity -> idiomatic. Another core contribution of our method is the use of Named Entity Recognition as an intermediate step to pre-identify some non-idiomatic expressions; this provides great improvements." \\
4&HIT& "1. we use the big pre-trained model, XLM-R-large. Compared with multilingual-BERT and XLM-R-base, XLM-R-large is obviously improved.
2. Separate the exact same phrases as MWE in the target sentence with the sep token. If the phrase in the sentence is capitalized, It is more likely to be named entities that the model can distinguish, so the sep tokens are not added around the capitalized phrases.
3. Using Regularized Dropout(r-drop) as regularization." \\
5&Hitachi& "Our approach is built on top of multilingual pre-trained language models, which include InfoXLM and XLM-R.
We solve the task of multilingual idiomaticity detection as a binary classification task and follow the standard fine-tuning method except not using a special [CLS] representation for classification. Instead, we first take an average over MWE's span representations and subsequently feed the averaged representation into a linear layer for classification." \\ 
6&OCHADAI& "our model relies on pre-trained contextual representations from different multilingual state-of-the-art transformer-based language models (i.e., multilingual BERT and XLM-RoBERTa), and on adversarial training, a  training method for further enhancing model generalization and robustness." \\
7&yjs& "For each input sentence in the training set, if the MWE is idiomatic then its corresponding tokens are labeled as "idiomatic" and the remaining tokens are labeled as "literal"; if the MWE is literal then all the tokens in the sequence are labeled as "literal". Method 1: We apply a Bi-Directional Attention Flow (BiDAF) network (Seo et al., 2017), while we use mBERT as the contextualised embedding, and we use pos tag embedding as its query input." \\
8&CardiffNLP-m& "CardiffNLP-metaphors submitted the results of two methods in total, applied both for Task A Zero Shot and one-shot. The first method uses xlm-roberta-large and the second uses several monolingual bert language models for English, Portuguese and Galician. For the Zero Shot settings, bert-multilingual-base is used to label the Galician sentences, because no Galician examples were included in the training set. The embedding of the three sentences and the embeddings of the isolated target are input of the models. We optimized the models over different training parameters on the development set." \\
9&Mirs& - \\
10&Amobee& - \\
11&HYU& "We devise four features ((i), (ii), (iii), and (iv) in the following) as input for a simple yet effective idiomaticity classifier that is a multi-layer perceptron with one hidden layer.

First, to consider the contextualized semantics of a target sentence when influenced by its surrounding context, we concatenate the target sentence with its (i) previous and (ii) next sentences respectively and inject the two chunks into our feature extractor (XLM-R; a bidirectional multilingual language model) independently to generate two distinct ((i) and (ii)) features.

While constructing the aforementioned features, we also introduce two techniques to clarify the presence of a MWE in the sequence: the first highlights the location of the MWE with a new, dedicated positional encoding, and the second appends the MWE once again at the end of the sequence.

In addition, we focus on the way of better utilizing the information existing solely in the target sentence, regarding a MWE and its context (i.e., phrases in the target sentence except for the MWE) as separate ones.

Specifically, we derive (iii) the “context-only” representation of the target sentence by using a variant of the target sentence where the MWE is masked, while we compute (iv) the “MWE-only” representation, which corresponds to the intrinsic meaning of the MWE irrespective of context, by inserting only the MWE into the feature extractor." \\
12&Zhichun Road& "1. We use InfoXLM-Base as text encoder. (performance: infoxlm > XLM-R > Mbert)
2.We use exponential moving average (EMA) method.
3.We use adversarial attack strategy(performance: Smart > freeLb > PGD = FGM).
Finally，our approach ranked 12th." \\ 
13&海鲛NLP& - \\
14&UAlberta& "Our unsupervised translation-based approach leverages translation information in multilingual resources such as OMW and BabelNet. The hypothesis is that the translations of idiomatic MWEs tend to be non-compositional, and therefore the individual words of an MWE are less likely to share mult-synsets with their translations. In addition, since MWEs that are named entities are usually literal, we use a part-of-speech tagger to identify proper nouns." \\ 
15&Helsinki-NLP& "The system utilizes linguistically motivated features that typically characterize idiomatic expressions: non-substitutability, non-compositionality and affectiveness. This feature model is based on pre-trained models and classification pipelines that have been integrated into the transformers library provided by HuggingFace. The final classification combines the feature model with either sentence-transformers or a base BERT model. The system also adds a back-translation feature and applies simple post-correction rules based on boolean features." \\ 
16&daminglu123& "We used the same model as baseline but added one more LSTM layer at last." \\ 
17&kpfriends & "We experimented with various inductive training methods only using Zero Shot data provided. 
We are still experimenting various schemes, including novel MWE ideas.
We will share the findings in our paper." \\ 
18&Unimelb\textunderscore AIP& "We tackled this task in an unsupervised way (i.e. without using any portion of the training data). First, we trained a standard CBOW word2vec model on unlabelled data and used it to predict the top-500 words that would fit into the surrounding context of the target MWE (as performed during the training of the CBOW model). Then, we calculated the maximum cosine similarities between the predicted words and each MWE component word, and regarded the MWE as “literal” (“non-idiomatic”) if they are higher than the mean cosine similarity between the component words and their 500 closest words. Finally, we ensembled five CBOW models trained with different window sizes (5, 10, 15, 20, and 30) to incorporate different levels of contextual information. One limitation of this approach is that it often classifies proper-noun and idiomatic usages into the same class (“non-literal”; as their surrounding contexts differ a lot from the literal usage ones), and to mitigate this problem, we always regarded MWEs as “non-idiomatic” if they contained any capital letter." \\ 
19&YNU-HPCC& "As for methods of the best submission results, we added a linear layer so as to choose effective information from all of output layer that were extracted by pre-trained model, XLM-RoBERTa, and then fine-tuned it to classify." \\ 
20&Ryan Wang& "CNN-bidirectional LSTM classifier with jointly trained word embeddings trained on full passages (target and context) from Zero Shot data" \\
\hline
N/A& JARVix & "we fine-tune a pretrained XLNet on the task dataset (after evaluating multiple large language models and their majority-voting ensemble)." \\
\hline
\end{tabularx}
\caption{\label{table:subtask-a-zero-shot-methods} Methods used in Subtask A Zero Shot. Note: CardiffNLP-m is short for CardiffNLP-metaphors.}
\end{table*}
\end{CJK*}

\begin{CJK*}{UTF8}{gbsn}

\begin{table*}[ht]
\tiny
\def\arraystretch{1.1}
\centering
\begin{tabularx}{\linewidth}{|c|c|p{12.75cm}|}
\hline
Ranking & Team & Method \\
\hline
1&HIT&"Mostly the same as the zero-shot. We train the One Shot model initialized from the best Zero Shot checkpoint. We additionally post-processed the predictions based on the distribution of the labels in the One Shot train file." \\
2&kpfriends& "More than 10 checkpoints were created per “English” and “Spanish / Galician” and inferred separately, later ensembled using soft-voting.
To stabilize training of xlm-roberta-large, we started with pre-trained models provided by Huggingface which were xlm-roberta-large trained on CoNLL.
We also had some good results with xlm-roberta-base.
We will deep dive into methodology and interesting observations / error analysis in our paper." \\ 
3&UAlberta& "Our method uses a transformer-based sequence classifier that takes as an input the context sentence and the glosses of each individual word in the target multi-word expression. The intuition is that the addition of the glosses to the input might help the classifier to detect if the meaning of the target multi-word expression can be deduced from the definitions of the individual words, i.e., if it is compositional. Note that this method is applicable to both settings." \\
4&Zhichun Road& "1. We use InfoXLM-Base as text encoder. (performance: infoxlm > XLM-R > Mbert)
2.We use exponential moving average (EMA) method.
3.We use adversarial attack strategy(performance: freeLB > Smart > PGD = FGM).
Finally，our approach ranked 4th." \\
5&clay& "same as Zero Shot setting, but with more data include Zero Shot data and One Shot data" \\
6&YNU-HPCC& "As for methods of the best submission results, we simply concated sentence and MWE and input into pre-trained model, XLM-RoBERTa.  CLS from last layer was extracted to classify." \\
7&CardiffNLP-m& "CardiffNLP-metaphors submitted the results of two methods in total, applied both for Task A Zero Shot and one-shot. The first method uses xlm-roberta-large and the second uses several monolingual bert language models for English, Portuguese and Galician. For the Zero Shot settings, bert-multilingual-base is used to label the Galician sentences, because no Galician examples were included in the training set. The embedding of the three sentences and the embeddings of the isolated target are input of the models. We optimized the models over different training parameters on the development set. xlm-roberta-large significantly ouperforms the monolingual experimental settings on the one shot track. " \\
8&yxb& "use mT5-Base
use Easy Data Augmentation techniques include back-translation, synonym replacement, random insertion, random swap, random deletion
include label unbalanced loss function：focal loss
use model ensemble" \\
9&NER4ID& "Same as zero-shot" \\
10&HYU& "In One Shot setting, we used the same method as in a Zero Shot setting." \\ 
11&yjs& "Method 2: We used the BiDAF-based DISC architecture by (Zeng and Bhat, 2021). DISC firstly combine GLOVE embeddings and POS embeddings with a BiDAF layer, which is then infused with mBERT by another BiDAF layer. 
We use both methods in the two settings, Method 1 performs better than Method 2.
In the submissions, the different results is caused by different random seeds, with/without previous and next sentences, and with/without MWE." \\ 
12&Mirs& - \\ 
13&daminglu123& "We used the same model as baseline but added one more LSTM layer at last." \\
14&海鲛NLP& - \\ 
15&OCHADAI& "our model relies on pre-trained contextual representations from different multilingual state-of-the-art transformer-based language models (i.e., multilingual BERT and XLM-RoBERTa), and on adversarial training, a  training method for further enhancing model generalization and robustness." \\ 
16&Ryan Wang& "CNN-bidirectional LSTM classifier with jointly trained word embeddings trained on full passages (target and context) from zero- and One Shot data" \\ 
\hline
N/A&MaChAmp& "Multi-task learning across SemEval tasks (2, 3, 4, 6, 10, 11, and 12). First we Pre Train a RemBERT multi-task model across all the tasks. Then we re-train a model for each task specifically. We used the default hyperparameters of MaChAmp v0.3 for all settings, which were finetuned on the GLUE benchmark and UD\textunderscore English-EWT." \\ 
N/A& JARVix & "we use a relation network (Sung, et. al 2018) to find a similarity (or a dissimilarity) score between a query and it's same MWE support set, and assign a label accordingly. For this, we also evaluate a siamese network with a similar inference methodology." \\
\hline
\end{tabularx}
\caption{\label{table:subtask-a-one-shot-methods} Methods for Subtask A One Shot. Note: CardiffNLP-m is short for CardiffNLP-metaphors.}
\end{table*}
\end{CJK*}

\begin{table*}[ht]
\tiny
\def\arraystretch{1.1}
\centering
\begin{tabularx}{\linewidth}{|c|c|p{12.8cm}|}
\hline
 Ranking & Team & Method  \\
\hline
1 & drsphelps & "Our model is a modification of the baseline system with the randomly initialised word embeddings for the one token MWEs replaced with embeddings created using Schick and Schutze's BERT for Attentive Mimicking (BERTRAM). BERTRAM models are trained for Portuguese and Galician alongside the provided English model, and examples use to create the MWE emebddings are taken from the common crawl corpora for English, Portuguese, and Galician. Further pretraining (up to 45 epochs) is done on the sentence transformers."\\
2 & colorful & - \\
3 & Mirs & - \\
4 & Zhichun Road & "1.We add CrossAttention-Module at the top of the Sentence-Bert. ( Including train and evaluate). 
2.We add an extra Contrastive Loss.
Finally, our approach ranked 4th." \\
5 & YNU-HPCC & "As for methods of the best submission results,  we extracted first-last-average vector and used  an optimized method called CoSENT to train model. In comparison to SBERT,  it could solve  the problem of difference in process of training and prediction and get a better results." \\
6 & ALTA & - \\
\hline
\end{tabularx}
\caption{\label{table:subtask-b-pre-train-methods} Methods for Subtask B Pre Train.}
\end{table*}

\begin{table*}[ht]
\tiny
\def\arraystretch{1.1}
\centering
\begin{tabularx}{\linewidth}{|c|c|p{12.8cm}|}
\hline
 Ranking & Team & Method \\
\hline
1 & YNU-HPCC & "As for methods of the best submission results, both multiple-negatives-ranking-loss and triplet-loss function combined with pre-trained model, distiluse-base-multilingual-cased-v1, were used to fine-tune. " \\
2 & drsphelps & "Using the models trained for the Pre Train setting, fine-tuning is performed using the provided training data, just as in the baseline system. The best overall performance is found after fine tuning for one epoch, however training for up to 50 epochs can drastically increase Spearman Rank scores for the idiom only data, while causing much less performance drop on the general STS data." \\
3 & Eat Fish & "Multilingual model which was pretrained by using knowledge distillation
Data augmentation
Extract multiword from exist multiword package
Two state training trick" \\
4 & Zhichun Road &"1.We add CrossAttention-Module at the top of the Sentence-Bert. ( Including train and evaluate). 
2.We add an extra Contrastive Loss.
Finally, our approach ranked 4th." \\
5 & ALTA & - \\
\hline
\end{tabularx}
\caption{\label{table:subtask-b-fine-tune-methods} Methods for Subtask B Fine Tune.}
\end{table*}

\end{document}